\documentclass[conference]{IEEEtran}
\IEEEoverridecommandlockouts
\usepackage{cite}
\usepackage{amsmath,amssymb,amsfonts}
\usepackage{graphicx}
\usepackage{textcomp}
\usepackage{xcolor}
\def\BibTeX{{\rm B\kern-.05em{\sc i\kern-.025em b}\kern-.08em
    T\kern-.1667em\lower.7ex\hbox{E}\kern-.125emX}}
    
\usepackage{fancyhdr}
\usepackage{xcolor,soul,framed}
\usepackage{xspace}
\usepackage{array}
\usepackage{eqparbox}
\usepackage{url}
\usepackage{longtable}
\usepackage{lipsum}
\usepackage{blindtext}
\usepackage{makecell}
\usepackage{mathtools}
\usepackage{commath}
\usepackage{multirow}
\usepackage{tabularx}
\usepackage[normalem]{ulem}
\usepackage{xspace}
\usepackage{algorithm}
\usepackage{algpseudocode}
\usepackage{amsfonts}
\usepackage{placeins}
\usepackage{amssymb}
\usepackage{pifont}

\usepackage[pagebackref=true,breaklinks=true,colorlinks,bookmarks=false]{hyperref}
\hypersetup{
    colorlinks=true,
    filecolor=magenta,      
    urlcolor=magenta,
}

\newcommand{\cmark}{\ding{51}}%
\newcommand{\xmark}{\ding{55}}%
\def\BibTeX{{\rm B\kern-.05em{\sc i\kern-.025em b}\kern-.08em
    T\kern-.1667em\lower.7ex\hbox{E}\kern-.125emX}}

\begin{document}

\title{Spatiotemporal Contrastive Learning of\\Facial Expressions in Videos}
\author{\IEEEauthorblockN{Shuvendu Roy, Ali Etemad}
\IEEEauthorblockA{Dept. ECE and Ingenuity Labs Research Institute \\
Queen's University, Kingston, Canada\\
\{shuvendu.roy, ali.etemad\}@queensu.ca}}

\maketitle
\thispagestyle{fancy}

\begin{abstract}
We propose a self-supervised contrastive learning approach for facial expression recognition (FER) in videos. We propose a novel temporal sampling-based augmentation scheme to be utilized in addition to standard spatial augmentations used for contrastive learning. Our proposed temporal augmentation scheme randomly picks from one of three temporal sampling techniques: (1) pure random sampling, (2) uniform sampling, and (3) sequential sampling. This is followed by a combination of up to three standard spatial augmentations. We then use a deep R(2+1)D network for FER, which we train in a self-supervised fashion based on the augmentations and subsequently fine-tune. Experiments are performed on the Oulu-CASIA dataset and the performance is compared to other works in FER. The results indicate that our method achieves an accuracy of 89.4\%, setting a new state-of-the-art by outperforming other works. Additional experiments and analysis confirm the considerable contribution of the proposed temporal augmentation versus the existing spatial ones.   
\end{abstract}

\begin{IEEEkeywords}
Self-Supervised Learning, Contrastive Learning, Facial Expressions, Affective Computing 
\end{IEEEkeywords}

\section{Introduction}

The objective of facial expression recognition (FER) is to predict basic expressions such as happy, disgust, neutral, sad, Surprised, and others, from facial images or videos \cite{pini2017modeling, yang2018facial}. Applications of such systems include emotion-aware multimedia and smart devices \cite{cho2019instant}, personal mood management \cite{thrasher2011mood, sanchez2013inferring}, and others. FER is challenging due to a number of reasons including variations in how different people express different emotions, different lighting or background conditions, and more. Moreover, the intensities with which emotions are expressed throughout the face can vary among subjects or even for the same subject in different contexts. 
For these reason, there has been an increased effort toward facial expression recognition in recent years from both images \cite{pini2017modeling, yang2018facial} and videos \cite{zhang2020facial, yang2018facial, ding2017facenet2expnet}.  

A variety of different machine learning and computer vision solutions have been proposed for FER \cite{li2020attention, liang2021patch}. More recently, however, due to the immense success of deep learning algorithms, deep neural networks (DNN) such as convolutional neural networks (CNN) have been successfully used to perform automatic feature extraction from training data. Nonetheless, deep networks generally need considerable amounts of `labeled' data to learn effective and robust features.

To reduce the reliance on output class labels during training, and also to learn better representations, self-supervised learning \cite{chen2019self, misra2020self, rahimi2020self} has emerged as an effective method to pre-train deep neural networks. Different approaches for self-supervised learning such as contrastive learning techniques like SimCLR \cite{chen2020simple} make use of the input data and create pseudo-labels using augmentations to train the network. Experiments have shown that such approaches can achieve competitive results even without the full datasets.

\begin{figure}[!t]
\centerline{\includegraphics[width=0.8\columnwidth]{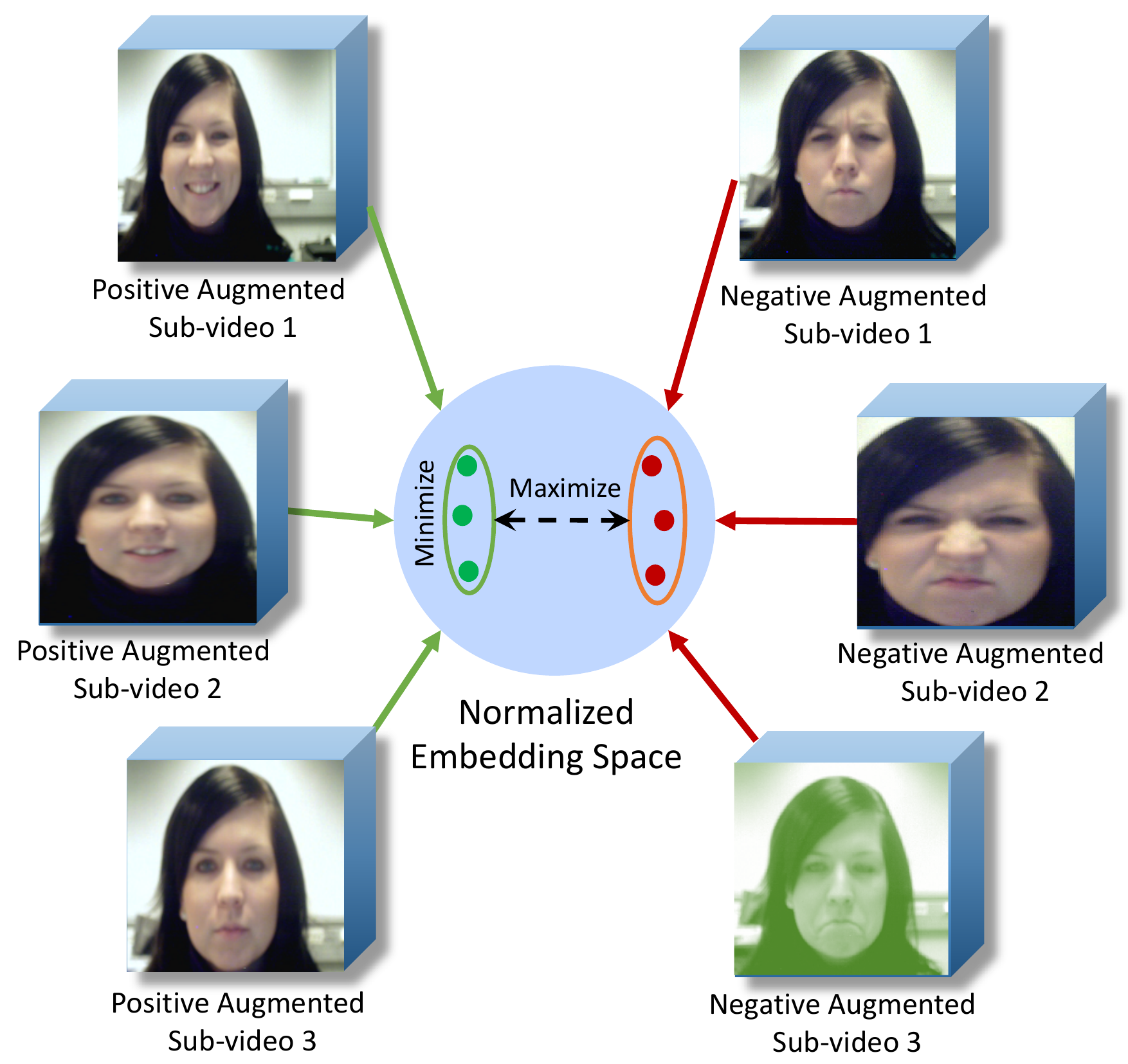}}
\caption{Contrastive self-supervised learning using various augmentations. Here augmented samples from the same video are considered positive examples, while other examples are considered negative ones.}
\label{fig_intro}
\end{figure}

In this paper, to take advantage of the desired properties of self-supervised learning, we propose a new video-based FER solution using contrastive learning. Our proposed method, Spatiotemporal Contrastive Learning of Representations (ST-CLR), first uses a novel temporal sampling scheme (temporal augmentation) to pick individual frames from the unlabeled input video to construct a sub-video. This will allow for various sub-videos to be constructed from any given input, resulting in better generalization. It then performs standard spatial augmentations, namely random cropping, color distortion, and random flipping to learn by maximizing the agreement between positive examples (augmentations on the same video) and minimizing the agreement for the negative pairs. A visual representation of our approach is shown in Figure \ref{fig_intro}. The figure demonstrates how the sub-video samples are projected onto a normalized embedding space, followed by minimizing of the distance between the positive pairs while maximizing the distance between the negative samples vs. the positives. We integrate this technique into a deep neural network that uses a R(2+1)dimensional framework instead of 3D convolutions to learn spatiotemporal representations. We perform extensive experiments and show the strong performance of our method on the Oulu-CASIA dataset \cite{taini2008facial}, outperforming all the previous methods.

In summary, our contributions are as follows.\\
\begin{itemize}
  \item We use contrastive self-supervised learning for FER in videos. To the best of our knowledge, this is the first time contrastive learning is used in this context.
  \item We introduce ST-CLR, an extension of the original SimCLR that performs temporal augmentations (sampling) followed by spatial augmentations for contrastive learning in videos.
  \item Using the proposed technique, we develop an end-to-end R(2+1)D network for video-based FER and perform rigorous experiments on the Oulu-CASIA dataset. The experiments show that our method outperforms existing works and sets a new \textit{state-of-the-art} on this dataset.
\end{itemize}

The rest of this paper is organized as follows. In the following section we review the related literature on self-supervised learning and video-based FER. This is followed by a detailed description of our proposed solution. Next, the experiment setup and implementation details are described, followed by the experimental results. Lastly in the final section, concluding remarks, limitations, and future work are discussed.

\section{Related Works}

\subsection{Self-Supervised Learning}

Self-supervised learning has been proposed to reduce the reliance of learning techniques on output labels such as human annotations, by applying specific augmentations \cite{chen2020simple} or transformations \cite{noroozi2017representation, dosovitskiy2015discriminative} to input data to generate pseudo-labels with which the model is trained. 
Training of the model using the generated pseudo-labels can be seen as an auxiliary pre-training task, also sometimes referred to as a 'pretext' task. Successive to the pre-training step by means of the pretext tasks, main components of the model, for instance the convolutional blocks in a CNN, are generally frozen and transfer learning is used to develop another model for 'down-stream' tasks, i.e. the main aim classification problem \cite{dosovitskiy2015discriminative}. In this phase, the fully connected layers are often trained from scratch with the frozen convolutional blocks using the original labels in the dataset \cite{chen2020simple}. In some cases, fine-tuning of the frozen components can also help achieve a better solution \cite{noroozi2016unsupervised}. This approach has been used on a variety of different problem domains including image analysis \cite{noroozi2016unsupervised, chen2019self}, wearable-based activity recognition \cite{saeed2019multi, rahimi2020self}, and affective computing with bio-signals \cite{sarkar2020self, sarkar2020detection, sarkar2020}.

In the context of image analysis, when trained with pretext tasks, the shallower layers of a CNN learn to extract low-level general purpose features such as corners, textures, and edges, whereas the deeper layers of the model learn high-level features such as detecting objects and larger parts of scenes \cite{bau2017network}. Recently, a number of different techniques have been proposed for self-supervised learning in images \cite{doersch2015unsupervised, noroozi2016unsupervised, zhang2016colorful, gidaris2018unsupervised, pathak2016context}. The novelties in most of these techniques lie in the way with which the pretext tasks are defined. For instance, in one approach, input images were divided into a number of blocks and shuffled, with the pretext task being to find the right order of blocks \cite{doersch2015unsupervised, noroozi2016unsupervised}. In \cite{gidaris2018unsupervised} images were distorted with different degrees of rotation, where the pretext task was to predict the amount of rotation that was applied to a distorted image. In \cite{zhang2016colorful} the pretext task was to colorize an input gray-scale image while in \cite{pathak2016context} the task was to fill missing parts of an image that was cropped out.

Recently contrastive learning has emerged as a powerful self-supervised learning technique and has shown great promise and achieved state-of-the-art results for a variety of different tasks \cite{hadsell2006dimensionality, dosovitskiy2014discriminative, oord2018representation, bachman2019learning}. In particular, a more recently proposed contrastive method called SimCLR \cite{chen2020simple} provides a generalized learning framework that does not requires any specialized architecture \cite{bachman2019learning, henaff2020data} nor any memory bank \cite{wu2018unsupervised, he2020momentum, misra2020self}. SimCLR uses data augmentation to obtain two representations of the same input which they consider as positive pairs. SimCLR learns the visual representation by maximizing the agreement of positive pairs via a contrastive loss in the latent space.

\subsection{Self-Supervised Learning in Video}

Self-supervised learning with video data often uses the temporal information in the video sequence, for instance tracking the temporal order of the frames of a video \cite{wang2015unsupervised, schroff2015facenet}. The intuition here is that the movement of objects in frames can be tracked in a sequence of consecutive video frames, thus the representation of frames that are temporally close in the video must be similar. In \cite{wang2015unsupervised}, a self-supervised method was proposed that takes a set of consecutive frames of a video to use as the main input. This was followed by selecting two additional sets from the same video, where the one temporally closest to the main set being designated as the positive set, while the other being designated as the negative set. This approach learned the video representations by enforcing the embeddings obtained from of the two close sub-videos to be similar in the latent space, while forcing the embedding of the negative pair to be far apart in the latent space. In \cite{schroff2015facenet}, the performance of \cite{wang2015unsupervised} was improved by picking the negative sub-video from a completely different video. In \cite{qian2020spatiotemporal}, a contrastive learning method was proposed for videos that, similar to \cite{wang2015unsupervised}, uses consecutive frames to form the positive and negative pairs.

As another approach to performing temporal self-supervised learning, a pretext task was designed to validate whether or not the frames of a video are in the right order successive to performing temporal shuffling \cite{misra2016shuffle}. Using this approach, the performance of video classification on large datasets, for instance for action recognition, was improved. Later, the Odd-One-Out Network \cite{fernando2017self} was proposed with a similar objective. In this case, given a few videos, the objective was to identity which video does not have the correct temporal order.

In order to not \textit{explicitly} learn sequence information, video colorization \cite{vondrick2018tracking} was proposed for self-supervised learning of videos. In this approach, the objective was to colorize input gray scale videos using a pre-colored reference frame. Unlike colorizing a single image, the temporal coherence in a video needed to be preserved, hence an loss function was designed to keep track of correlated pixels in consecutive frames.

\subsection{Expression Recognition from Video}

For sequence-based \cite{sepas2020facial, sepas2021capsfield} or video-based \cite{pini2017modeling, yang2018facial} FER, RNNs such as LSTM networks have been used to learn temporal information along with CNNs that exploit spatial information from the individual video frames \cite{lu2018multiple, sun2016lstm}. Alternatively, 3D CNNs have also been explored for video-based FER \cite{liu2014deeply, lu2018multiple}. For example, a 3D CNN was used with deformable action parts constraints for dynamic motion encoding in FER called CNN-DAP \cite{liu2014deeply}. In \cite{lu2018multiple}, a 3D CNN was used in an ensemble with two 2D CNN models. Unlike 3D CNNs that share weights in the time dimension, in \cite{jung2015joint}, a method was proposed that used a sequence of images without weight sharing in the time dimension.

Similar to the nature of self-supervised training, fully supervised pre-training with additional image or video datasets has been widely explored in the literature. In \cite{kuo2018compact}, a frame-to-sequence method was proposed that used a pre-trained CNN as a feature extractor followed by an RNN to capture the temporal features. In \cite{pini2017modeling}, a multi-modal framework was proposed that used both image and video -level information with an ensemble of 2D and 3D CNNs. Here the 2D model acts as a pre-trained feature extractor which is followed by a NetVLAD layer \cite{arandjelovic2016netvlad} which aggregates the temporal representation of videos. In \cite{yang2018facial}, a generative approach was proposed to pre-train the network  with a conditional generative adversarial network \cite{goodfellow2014generative}. 

Finally, another approach for enhancing performance in video-based FER has been to focus on particular parts of the video sequence where expressions are most prominent \cite{zhao2016peak}. The difficulty with this method is the lack of knowledge as to when the prominent expressions occur. To tackle this, attention mechanisms can be used to focus on salient parts of sequences, as performed in \cite{li2020attention, sepas2020facial, zhang2019classification}.

\begin{figure*}[htbp]
\centerline{\includegraphics[width=0.94\linewidth]{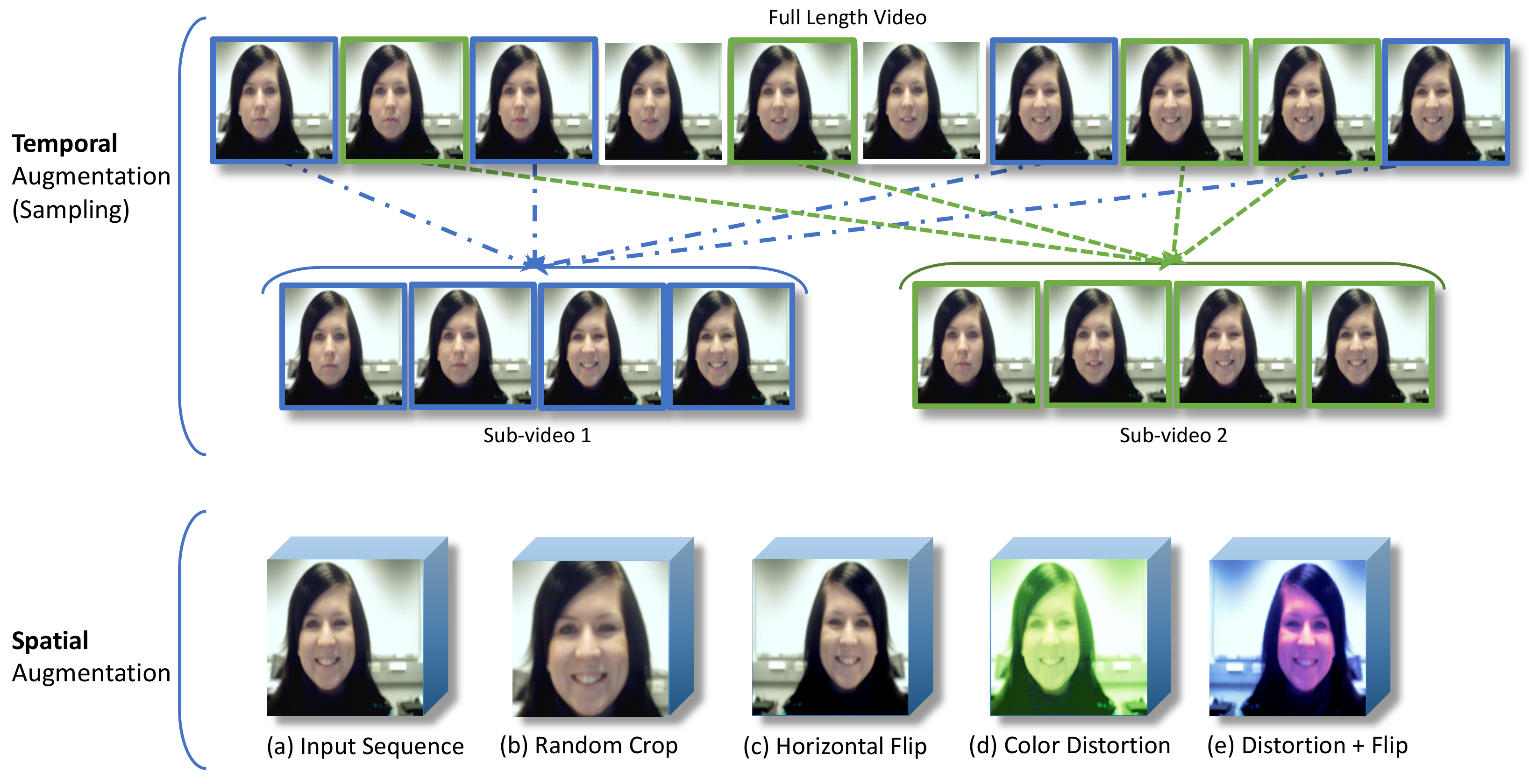}}
\caption{An example of the temporal and spatial augmentations used in ST-CLR are depicted.}
\label{fig_3}
\end{figure*}

\section{Method}
The proposed method is a 2 step training process. First, the model is pre-trained by contrastive self-supervised learning. Then the model is fine-tuned for the task of video-based FER. Following we present our method in detail. 

\subsection{Self-Supervised Training}
The self-supervised step in our method is inspired by the recent developments in contrastive learning, which aim to maximize the agreement between augmented versions of the same input in latent space using contrastive loss. The proposed method consists of three main components, namely stochastic sampling and augmentation, an encoder, and a projection head.

\subsubsection{Sampling and Augmentation Module}

We propose a temporal sampling step prior to spatial augmentation, which can also be viewed as a `\textit{temporal augmentation}' operation. This is followed by the standard image-based augmentations used in \cite{chen2020simple}. Accordingly, given input video $x$, we sample two sub-videos $\Tilde{x}_1$ and $\Tilde{x}_2$. Next, augmentations are performed on each sub-video to yield $\Tilde{x}^{Aug}_1$ and $\Tilde{x}^{Aug}_2$. The augmented sub-videos that are sampled from the same video are considered positive examples, while any augmented sub-video sampled from a different video is considered a negative example. Following we describe the two augmentation steps (temporal and spatial).

\textbf{Temporal Augmentation}. 
To sample a given input video, we defined 3 sampling strategies: 
(1) \textit{Pure random sampling} where $n$ frames are sampled at random from the full length video, then sorted according to the order in the original video; 
(2) \textit{Uniform sampling} where frames are sampled uniformly starting and ending with the first and last frames of the input while maintaining uniform distance in between consecutive sampled frames; 
(3) \textit{Sequential sampling} where an initial frame is randomly picked, followed by the following sequence of $n-1$ frames. 
To perform the temporal augmentation step in our method, we randomly choose between the above strategies every time. Therefore, as we sample the sub-videos in a different way at each iteration, the positive pairs not only see different spatial augmentations of the same input video, but also sees a completely different sub-video. This results in better generalization and robustness to temporal variations. Figure \ref{fig_3} (top) depicts the temporal augmentation step where two sub-videos are sampled using the first sampling strategy.
    
\textbf{Spatial Augmentation}. Successive to temporal sampling, we apply the following augmentations to the entire sub-video sequences to obtain $\Tilde{x}^{Aug}_i$.
(1) \textit{Random resized crop}. First, random cropping is applied to a given sub-video. Here, a frame size and position is picked at random, then all the frames of that sub-video are cropped at the same position. Finally, the video sequence is resized to the desired input shape of the CNN model.
(2) \textit{Random flip}. Finally, we randomly choose whether to horizontally flip all the frames in a sub-video or not. 
(3) \textit{Random color distortion}. This augmentation includes modifying brightness, contrast, saturation, and hue of the sub-videos. Here, the distortion parameters are picked randomly and applied to all the frames of the sub-video.
It should be noted that unlike the temporal augmentation step where only one of the sampling strategies is picked for each sub-video, for spatial augmentation, all three operations are performed together. Nonetheless, as the parameters are chosen at random, zeros could be chosen for the parameters of a certain spatial augmentation, effectively not perform it. Figure \ref{fig_3} (bottom) shows examples of spatial augmentation operations performed on a sample frame.

\begin{figure}[!t]
\centerline{\includegraphics[width=0.6\columnwidth]{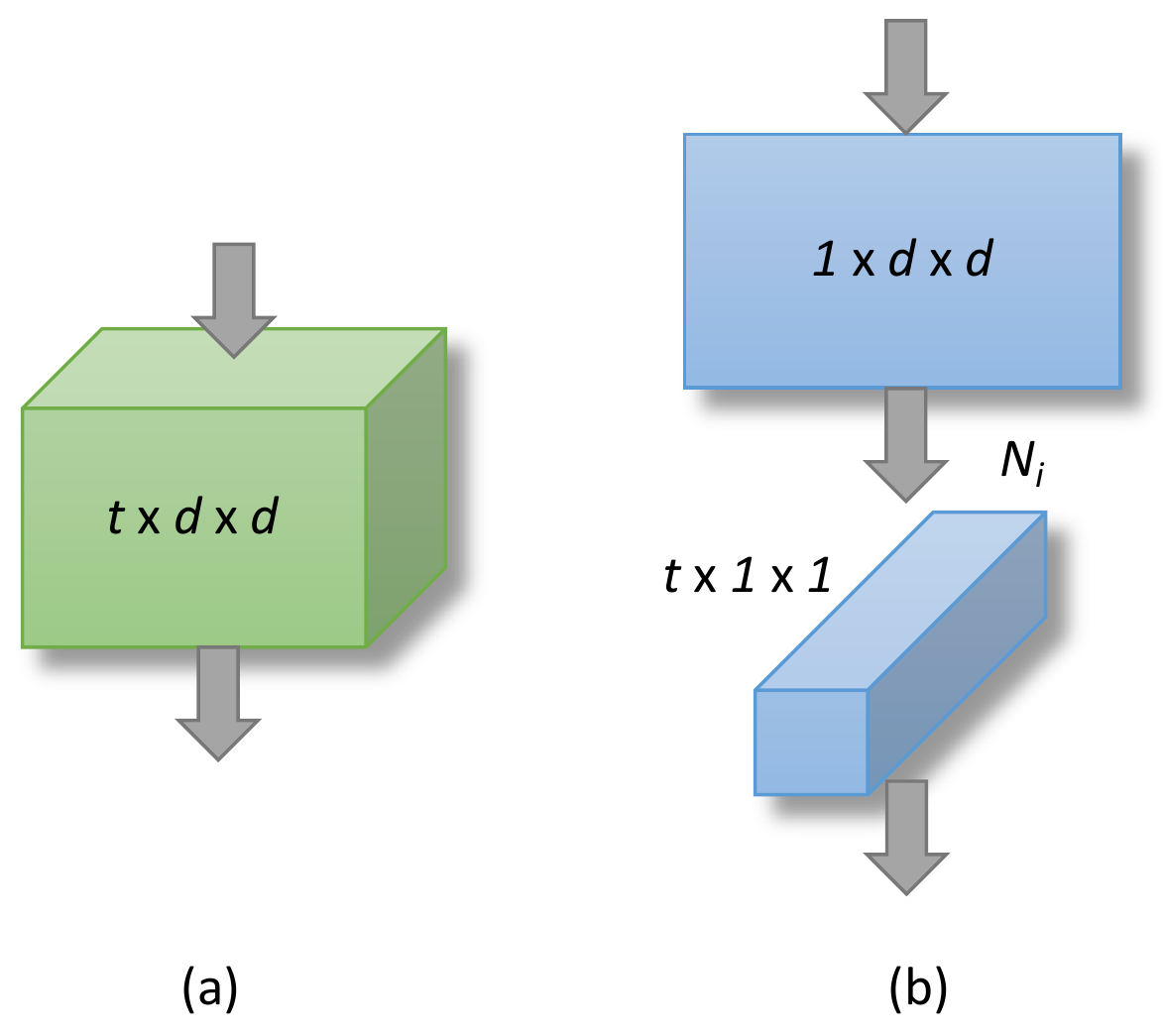}}
\caption{Factorization of 3D convolutional block into (2+1)D.}
\label{fig_2plus1}
\end{figure}

\begin{figure}[!t]
\centerline{\includegraphics[width=0.4\columnwidth]{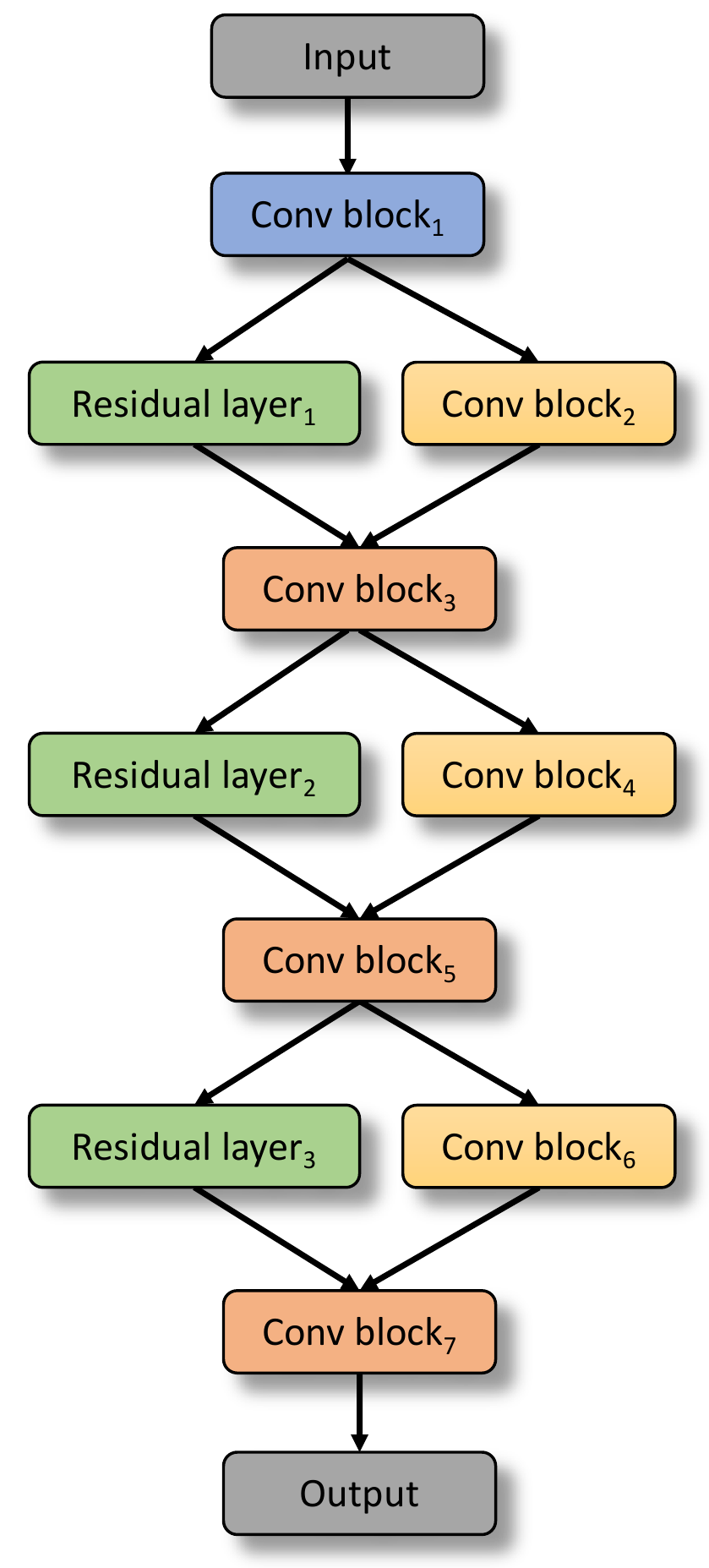}}
\caption{The R(2+1)D encoder network of the proposed method.}
\label{fig_6}
\end{figure}

\subsubsection{Encoder Module $f(x)$} 

We use an encoder block (a simple CNN) to transform the input video into a latent representation. A very common way for modeling 4-dimensional data such as videos is using 3D convolutional neural networks \cite{tran2015learning, liu2014deeply, lu2018multiple}. This strategy allows for both spatial and temporal dimensions to be directly learned. However, 3D CNNs have a large number of parameters, requiring significantly more data and processing power to train. To address this drawback,  an approach was proposed to factorize the 3D CNNs \cite{tran2018closer}, breaking down the 3D convolutions into a 2D operation followed by a 1D operation. The resulting model with residual connections \cite{he2016deep} was called R(2+1)D \cite{tran2018closer}. There are a couple of advantages to this model. First, the factorized operations contain a non-linear function in between the layers, resulting in more non-linearity in the network, which is a desired property. 
Second, the factorization helps the optimizer during training by lowering the training loss and training time. It was also shown in \cite{tran2018closer} that factorizing the 3D convolution into two distinct spatial and temporal operations improves the performance of learning in large scale video classification tasks. In our paper, we use this concept in the encoder block of the proposed network. This notion is depicted in Figure \ref{fig_2plus1}.

\begin{table}[htbp]
\caption{Architecture of Encoder module}
\begin{center}
\begin{tabular}{|c|cc|c|}
\hline
\textbf{Layer Name} &\textbf{ Output size}& &\textbf{ Kernel size} \\
\hline\hline
\multirow{4}{6.
em}{Conv Block$_1$}& [45, 16, 112, 112]  && 45, [1, 7, 7]\\

 & [64, 16, 112, 112]  && 64, [3, 1, 1] \\

 & [144, 16, 112, 112]  & \multirow{2}{3em}{\Big\} $\times 4$} & 144, [1, 3, 3] \\
                             & [64, 16, 112, 112]  &  & 144, [3, 1, 1] \\

\hline

\multirow{4}{6em}{Conv Block$_2$} & [230, 16, 56, 56]  &  & 230, [1, 3, 3] \\
& [128, 8, 56, 56]  &  & 128, [3, 1, 1] \\

 & [230, 8, 56, 56]  &  & 230, [1, 3, 3]\\
& [128, 8, 56, 56]  &  & 128, [3, 1, 1]\\
\hline

Residual layer$_1$ & [128, 8, 56, 56] && 128, [1, 1, 1] \\

\hline

\multirow{2}{6em}{Conv Block$_3$} & [288, 8, 56, 56]  & \multirow{2}{3em}{\Big\} $\times 2$} & 288, [1, 3, 3] \\
& [128, 8, 56, 56]  &  & 128, [3, 1, 1] \\

\hline

\multirow{4}{6em}{Conv Block$_4$} & [460, 8, 28, 28]  &  & 460, [1, 3, 3] \\
& [256, 4, 28, 28]  &  & 256, [3, 1, 1]\\

 & [460, 4, 28, 28]  &  & 460, [1, 3, 3] \\
& [256, 4, 28, 28]  &  & 256, [3, 1, 1] \\
\hline

Residual layer$_2$ & [256, 4, 28, 28] && 256, [1, 1, 1] \\

\hline

\multirow{2}{6em}{Conv Block$_5$} & [576, 4, 28, 28]  & \multirow{2}{3em}{\Big\} $\times 2$} & 576, [1, 3, 3] \\
& [256, 4, 28, 28]  &  & 256, [3, 1, 1] \\

\hline

\multirow{4}{6em}{Conv Block$_6$} & [921, 4, 14, 14]  &  & 921, [1, 3, 3]\\
& [512, 2, 14, 14]  &  & 512, [3, 1, 1] \\

 & [921, 2, 14, 14]  &  & 921, [1, 3, 3]\\
& [512, 2, 14, 14]  &  & 512, [3, 1, 1] \\

\hline

Residual layer$_3$ & [512, 2, 14, 14] && 512, [1, 1, 1] \\

\hline

\multirow{2}{6em}{Conv Block$_7$} & [1152, 2, 14, 14]  & \multirow{2}{3em}{\Big\} $\times 2$} & 1152, [1, 3, 3] \\
& [512, 2, 14, 14]  &  & 512, [3, 1, 1]\\

\hline

Ada. Ave. Pool & [512, 1, 1, 1]&&\\

\hline
\end{tabular}
\label{tab1}
\end{center}
\end{table}

\begin{figure*}[!t]
\centerline{\includegraphics[width=0.82\linewidth]{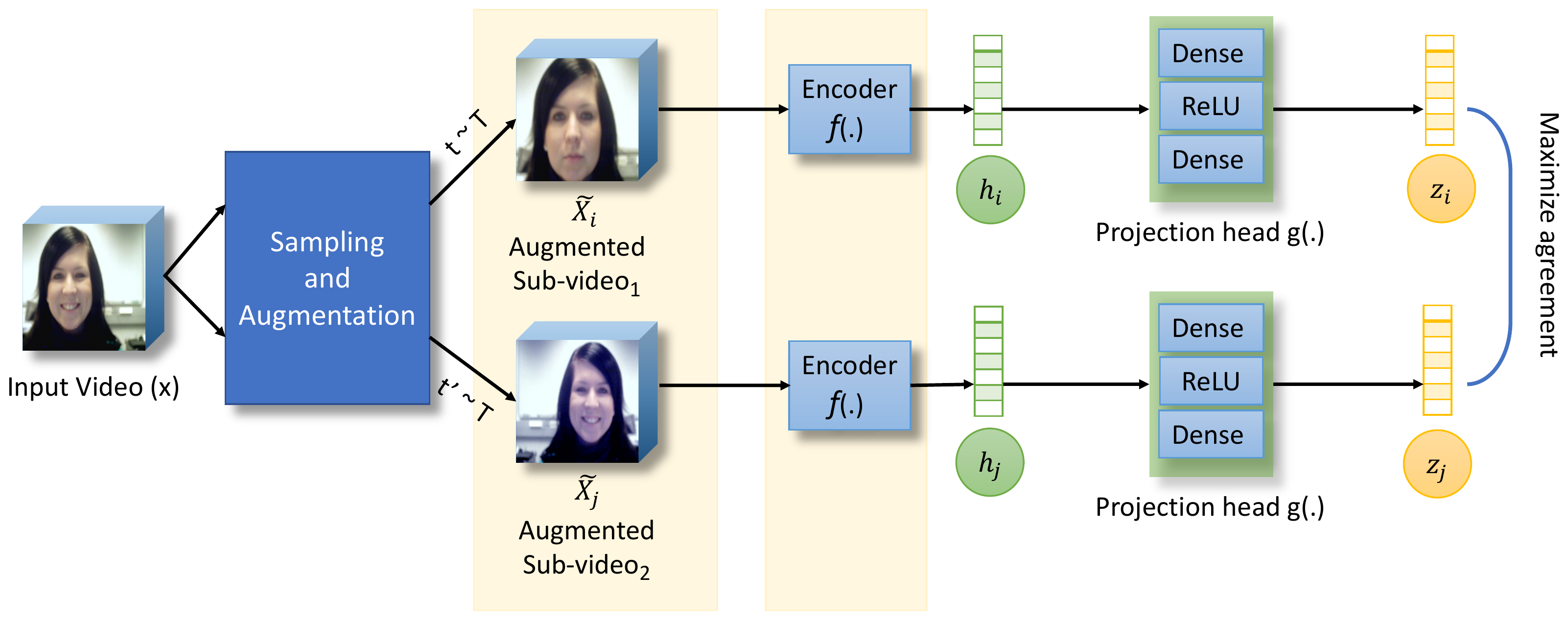}}
\caption{Visual illustration of our proposed method, where the agreement between different augmented sub-videos is maximized in the latent space.}
\label{fig_model}
\end{figure*}

To form a (2+1)D block from a 3D block of shape $M_{i-1}\times t \times d \times d$, where $M_{i-1}$ is the number of 3D kernels with each kernel size being $t \times d \times d$, we replace the $M_i$ 3D Convolutional kernels with (2+1)D blocks which contain $N_i$ 2D convolutional kernels of shape $M_{i-1} \times 1 \times d \times d$ and $M_i$ temporal convolutional kernels of shape $N_i \times t \times 1 \times 1 $. Here, $N_i$ is the dimension of the intermediate representations as shown in Figure \ref{fig_2plus1}. The encoder with all the CNNs and residual connections are shown in Figure \ref{fig_6} and the full architecture of the encoder model using the (2+1)D blocks is laid out in Table \ref{tab1}.  Lastly, we represent the output embedding of the encoder block as $h_i = f(\Tilde{x}_i)$, where $h_i \in \mathbb{R} ^d$. 

\subsubsection{Projection head $g(x)$} 
Following the intuition of \cite{chen2020simple}, we use a projection head $g(x)$ consisting of 2 dense layers, with a ReLU non-linearity in between, which takes the output of $f(x)$ as input and outputs the final latent embedding. Experiments in \cite{chen2020simple} showed improvements in the learnt representation by including such a non-linear learnable transformation. The final embedding is represented as $z_i = g(h_i)$ on which the contrastive loss is applied. 

\subsubsection{Contrastive loss} We use a contrastive loss to train the proposed self-supervised network. The loss is designed to learn visual representation from seeing positive and negative pairs. Let's define the similarity between the embedding $z_i$ and $z_j$ (the outputs from the encoder) as $cosine(z_i,z_j) = z_i^Tz_j/||z_i||.||z_j||$. If $i$ and $j$ are positive examples, then the contrastive loss function is defined as:
\begin{align}
Loss_{i, j} = -log\frac{exp(cosine(z_i, z_j)/\tau)}{\sum_{k=1}^{2N} \mathbf{1}_{[k \neq i]} exp(cosine(z_i, z_k)/\tau)},
\end{align}
where $\mathbf{1}_{[k \neq i]} \in \{0, 1\}$ is a function of $i$, $k$ which outputs 1 when $i \neq k$. Here $\tau$ represents a temperature parameter which scales the cosine similarity. In some previous works such as \cite{chen2020simple}, loss function defined above is termed normalized temperature-scaled cross entropy loss. Figure \ref{fig_model} depicts a visual illustration of the proposed ST-CLR method.

\subsection{Model Fine-tuning}
After pre-training the model with self-supervised learning, we no longer use the projection head at the end of the encoder block. During the supervised fine-tuning phase, we drop the projection head and add a single-layer linear transformation for generating the output probability. This fine-tuning is performed with categorical cross-entropy loss.

\section{Experiments and Results}

\subsection{Experiments}
In this section we present the experiments and the results for the proposed method on video-based FER. We conduct  the experiments on a popular video FER dataset called Oulu-CASIA \cite{taini2008facial}.

\textbf{Dataset.} The Oulu-CASIA \cite{taini2008facial} is a facial expression recognition dataset that contains 6 expressions: Happy, Sad, Surprised, Angry, Fear, and Disgust. This dataset was collected from 80 people aging from 23 to 58. Among them 73.8\% were male and 26.2\% were female. The dataset was collected in a lab environment where the subjects were asked to sit about 60 \textit{cm} away from the camera. The videos were collected in 25 frames per second at a spatial resolution of $320\times 240$.

\textbf{Implementation Details}
As mentioned in the method section, we use the R(2+1)D architecture for the encoder of our method. At the end of the encoder we have a projection head, which is a 2-layer network. The output of the projection head is a 128 dimensional vector, on which the contrastive loss is calculated. The model is trained with an SGD optimizer for 1000 epochs with a momentum of 0.9, a learning rate of 0.001, and weight decay of 1e-4. Oue solution was implemented in PyTorch and trained using 4 NVIDIA V100 GPUs. 
The model takes sub-videos with 16 frames, and spatial resolution of $224 \times 224$ as its inputs.

In the fine-tuning step, we train the final model for 100 epochs. First, we train the newly initialized linear layer for 30 epoch and then fine-tune the entire model for 70 epochs. The model is trained using Adam optimizer with a learning rate of 1e-4 and plateau learning rate decay with a patience of 3.

\subsection{Results}
\begin{figure}[!t]
\centerline{\includegraphics[width=.85\columnwidth]{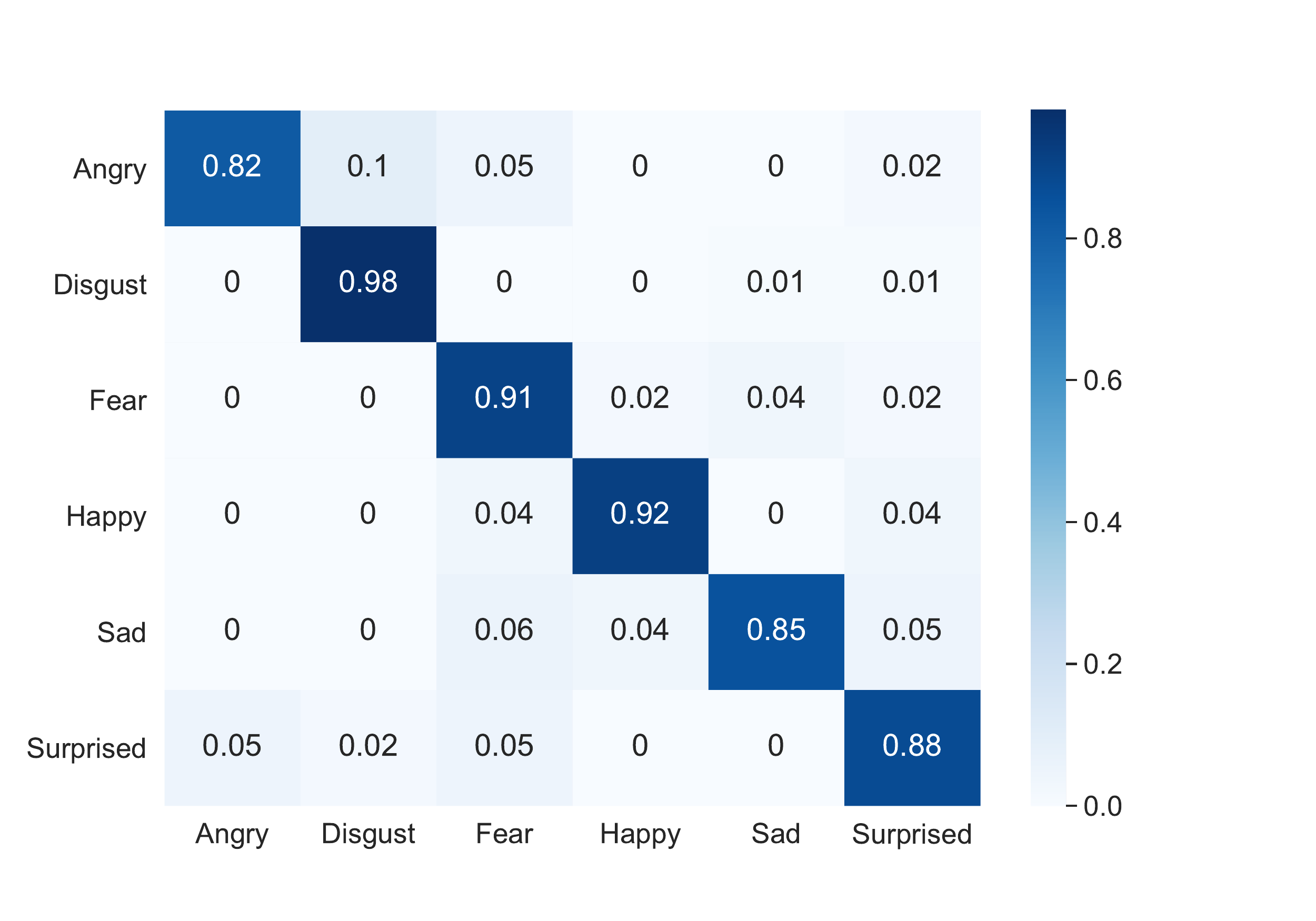}}
\caption{Confusion matrix of the prediction of the model on Oulu-CASIA dataset. Model predictions are presented in the $x$ axis while the ground-truths are in the $y$ axis.}
\label{fig_matrix}
\end{figure}

We train the proposed method on Oulu-CASIA dataset. To compare our method with previous works on this dataset, we have conducted all the experiments with 10-fold cross-validation. Table \ref{tab_result} compares the result of our solution with previous methods on the Oulu-CASIA dataset. Our ST-CLR method performs the best among all the methods in this comparison. The confusion matrix depicted in Figure \ref{fig_matrix} shows that our method performs the best on classifying `disgust' and also performs well on expressions `fear' and `happy'. We observe the least effective performance on `angry' and `sad'. The figure shows that `Angry' is often confused with `Disgust' while `Sad' is generally confused with `Fear' and `Surprised'. Lastly, `Surprised' is mostly confused with `Fear' and `Angry'. This analysis points to relatively reasonable mistakes by the model.

\begin{table}[!t]
\caption{Comparison with previous methods on Oulu-CASIA dataset}
\begin{center}

\begin{tabular}{l l c}
\hline
\textbf{Authors} & \textbf{Method} & \textbf{Accuracy}\\
\hline\hline
Zhao et al. \cite{zhao2007dynamic} & LBP-TOP & 68.13\% \\
Liu et al. \cite{liu2014learning} & STM-ExpLet & 74.59\% \\
Guo et al. \cite{guo2012dynamic} & Atlases & 75.52\% \\
Jung et al. \cite{jung2015joint} & DTAGN-Joint & 81.46\% \\
Zhao  et al. \cite{zhao2016peak} & PPDN & 84.59\% \\
Yu et al. \cite{yu2018deeper} & DCPN & 86.23\% \\
Zhangh et al. \cite{zhang2017facial} & PHRNN-MSCNN & 86.25\% \\
Ding et al. \cite{ding2017facenet2expnet} & FN2EN & 87.71\% \\
Yang et al. \cite{yang2018facial} & DeRL & 88.00\% \\
Zhang et al. \cite{zhang2020facial}& WMCNN-LSTM & 88.00\%\\
\hline
\textbf{Ours} & \textbf{ST-CLR} & \textbf{89.38}\%\\
\hline
\end{tabular}
\label{tab_result}
\end{center}
\end{table}

Next, we conduct ablation studies on the main elements of the proposed method. As discussed earlier the proposed method requires a CNN backbone as the encoder suitable for 4 dimensional input video data, for which we used R(2+1)D. We conduct experiments by replacing this backbone with two popular alternative encoder architectures, namely a CNN with 3D convolutions and a CNN with mixed 2D and 3D convolutions. For the 3D convolutional network, we replace all the R(2+1)D blocks mentioned in Table \ref{tab1} with 3D blocks. As for the mixed 3D-2D alternative, 3D blocks are used for all the convolutions in the $Conv Block_1$ and 2D blocks are used for rest of the network. Table \ref{tab_cnn} shows the accuracy of our model versus its variants when the alternative networks are used as the encoder block. As we can see, R(2+1)D performs better than other networks, confirming our choice of using the R(2+1)D architecture.

\begin{table}[!t]
\caption{Comparison of different encoder blocks on Oulu-Casia dataset}
\begin{center}

\begin{tabular}{l l}
\hline
\textbf{Encoder Network} & \textbf{Accuracy}\\
\hline\hline
3D Convolution & 87.50\%\\
Mixed 2D and 3D Convolution & 85.21\%\\
R(2+1)D Convolution & 89.38\%\\
\hline
\end{tabular}

\label{tab_cnn}
\end{center}
\end{table}

\begin{table}[!t]
\caption{Effect of different augmentations on Oulu-Casia dataset. FS: Frame Sampling, RRC: Random Resized Crop, RCD: Random Color Distortion, RF: Random Flip.}
\begin{center}
\begin{tabular}{lcccc|c}
\hline
\textbf{Augmentations} & \textbf{FS} & \textbf{RRC} & \textbf{RCD} & \textbf{RF} & \textbf{Accuracy} \\
\hline\hline
All (temp. + sp.) & \cmark & \cmark & \cmark & \cmark &  89.38\%\\
w/o temp.   & \xmark & \cmark & \cmark & \cmark &  85.41\%\\
temp. + partial sp. 1   & \cmark & \xmark & \cmark & \cmark &  87.71\%\\
temp. + partial sp. 2   & \cmark & \cmark & \xmark & \cmark &  86.67\%\\
temp. + partial sp. 3   & \cmark & \cmark & \cmark & \xmark &  89.17\%\\
\hline
\end{tabular}
\label{tab_aug}
\end{center}
\end{table}

\begin{table}[!t]
\caption{Comparison of performance for different amount of ssl epochs on Oulu-Casia dataset}
\begin{center}
\begin{tabular}{l l}
\hline
\textbf{Epochs} & \textbf{Accuracy}\\
\hline\hline
100 & 87.71\%\\
500 & 89.17\%\\
1000 & 89.38\%\\
\hline
\end{tabular}
\label{tab_epoch}
\end{center}
\end{table}

\begin{table}[!t]
\caption{Comparison of performance of the method using different amount of supervised data on Oulu-CASIA dataset}
\begin{center}
\begin{tabular}{l l}
\hline
\textbf{Labeled Data (\%)} & \textbf{Accuracy}\\
\hline\hline
75\% & 87.50\%\\
50\% & 83.33\%\\
25\% & 77.08\%\\
10\% & 54.17\%\\
\hline
\end{tabular}
\label{tab_partial}
\end{center}
\end{table}

Another important component of contrastive self-supervised leaning is the use of augmentations. One of the novelties in our proposed method is the use of frame sampling -based temporal augmentations along with popular spatial augmentations. Table \ref{tab_aug} shows the ablation study on spatial and temporal augmentations used in our method, where the impact of each augmentation is investigated by systematic exclusion. We see that the temporal augmentation clearly has the highest impact on the final performance as removing this augmentation module reduces the accuracy of the model by about 4\%. Among the spatial augmentations, `random color distortion' shows the highest impact, which is in conformity with other works on contrastive learning \cite{chen2020simple}.

It was shown in previous works such as \cite{chen2020simple} that self-supervised learning usually benefits from longer training compared to fully supervised learning. In Table \ref{tab_epoch}, we show the impact of different number of epochs for pre-training our model. We find that the accuracy improves significantly when the number of epochs increases from 100 to 500. However, only slight improvements are achieved when training it for another 500 epochs (for a total of 1000 epochs). So, for our final model, we pre-train the model for 1000 epochs.

Finally, to show the robustness of our model, we conduct experiments with partially labeled data. Table \ref{tab_partial} shows the accuracy of the model with different portions of the dataset being labeled and used for fine-tuning. We find that, using 75\% and 50\% data drops the accuracy by only 2\% and 6\% respectively. Even with only 25\% of the data being labeled, the accuracy is better than few of the previous works shown in Table \ref{tab_result}. This indicates that our model is robust to having fewer labeled data, which can be attributed to the self-supervised training strategy used.

\section{Conclusion and Future Work}
This paper presents a self-supervised approach for learning facial expressions in video. Our method uses a R(2+1)D style architecture to avoid using 3D convolutions for faster and better performance. To perform self-supervised training for our model, we propose a novel temporal augmentation scheme along with standard spatial (frame-based) augmentations. We test our solution on the Oulu-CASIA dataset and perform rigorous experiments. Our experiments show that the proposed method outperforms existing work on this dataset to set a new state-of-the-art. 
Extensive experiments were conducted to show the importance of each of the utilized augmentations toward the final performance of the model, highlighting the importance of our proposed temporal augmentation. Additional experiments were performed which validated our choice of R(2+1)D convolutions versus 3D and hybrid approached. Lastly, our analysis shows the advantage and robustness of our self-supervised method when significant portions of output labels were not used for training.

For future work, the proposed temporal augmentation scheme can be used for contrastive self-supervised learning in other video-based representation learning tasks such as action recognition, pose estimation, object detection and tracking, and others. Accordingly, additional modifications may need to be made to the network architecture and hyper-parameters in order to obtain optimum performance in other domains and with other datasets.

\section*{Acknowledgements}
We would like to thank BMO Bank of Montreal and Mitacs for funding this research.

\bibliographystyle{IEEEtran}
\bibliography{IEEEabrv,ref}

\end{document}